\documentclass[runningheads]{llncs}
\usepackage{graphicx}
\usepackage{amsmath,amssymb} 
\usepackage{color}
\usepackage{times}
\usepackage{multirow}
\usepackage{subfigure}
\usepackage{mathrsfs}

\begin{document}
\title{Learning Spectral Transform Network on 3D Surface for Non-rigid Shape Analysis} 
\titlerunning{Spectral Transform Network}
\author{Ruixuan Yu \and Jian Sun\thanks{Corresponding author.} \and Huibin Li}
\authorrunning{R. Yu et al}
\institute{Xi'an Jiaotong University, Xi'an, 710049, China
\email{yuruixuan123@stu.xjtu.edu.cn,\{jiansun,huibinli\}@xjtu.edu.cn}}
\maketitle             

\begin{abstract}
Designing a network on 3D surface for non-rigid shape analysis is a challenging task. In this work, we propose a novel  spectral transform network on 3D surface to learn shape descriptors. The proposed network architecture consists of four stages: raw descriptor extraction, surface second-order pooling, mixture of power function-based spectral transform, and metric learning. The proposed network is simple and shallow. Quantitative experiments on challenging benchmarks show its effectiveness for non-rigid shape retrieval and classification, e.g., it achieved the highest accuracies on SHREC'14, 15 datasets as well as the ``range'' subset of SHREC'17 dataset.

\end{abstract}

\section{Introduction}
3D shape analysis has  become increasingly important with the advances of shape scanning and  processing techniques. Shape retrieval and classification are two fundamental tasks of 3D shape analysis, with diverse applications in archeology,  virtual reality, medical diagnosis, etc. 3D shapes generally include rigid shapes,  e.g., CAD models, and non-rigid shapes such as human surfaces with non-rigid deformations. 

A fundamental problem in non-rigid shape analysis is  shape representation. Traditional shape representation methods are mostly based on local artificial descriptors such as shape context~\cite{Shapecontext}, mesh\text{-}sift~\cite{Dirk,li2015towards}, spin images~\cite{Andrew}, etc., and they have shown effective performance especially for shape matching and recognition. These descriptors are further modeled as middle level shape descriptors by Bag-of-Words model~\cite{BOWVLAD1}, VLAD~\cite{VLAD}, etc., and then applied to shape classification and retrieval. 
For shapes with non-rigid deformations, the model in~\cite{elad2003on} generalize shape descriptors from Euclidean metrics to  non-Euclidean metrics. The spectral descriptors, which are built on spectral decomposition of Laplace\text{-}Beltrami operator defined on 3D surface, are popular in non-rigid shape representation. Typical spectral descriptors include diffusion distance~\cite{lafon2006data}, heat kernel signature~(HKS)~\cite{Jian}, wave kernel signature~(WKS)~\cite{MathieuWKS} and scale invariant heat kernel signature~(SIHKS)~\cite{MMB1}. In~\cite{Ioannis}, spectral descriptors of SIHKS and WKS using a Large Margin Nearest Neighbor~(LMNN) embedding achieved state-of-the-art results for non-rigid shape retrieval.   Spectral descriptors are commonly intrinsic and invariant to isometric deformations, therefore effective for non-rigid shape analysis. 

Recently, a promising  trend in non-rigid shape representation is the learning-based methods on 3D surface for tasks of non-rigid shape retrieval and classification.  Many learning-based methods take low-level shape descriptors as inputs and extract high-level descriptors by integrating over the entire shape. In the work of~\cite{Ioannis}, they first extract SIHKS and WKS, and then integrate them to form a global descriptor followed by LMNN embedding.  Global shape descriptors are learned by Long-Short Term Memory (LSTM) network in~\cite{YiFang1} based on spectral descriptors. The eigen-shape and Fisher-shape descriptors are learned by a modified auto-encoder based on spectral descriptors in~\cite{YiFang3}. 
These works have shown impressive results in learning global shape descriptors. Though these advances have been achieved, designing learning-based methods on 3D surface is still an emerging and challenging task, including how to design feature aggregation  and feature learning on 3D surface for non-rigid shape representation. 

\begin{figure*}[t]
\begin{center}
\includegraphics[width=1\linewidth,height=1.5in]{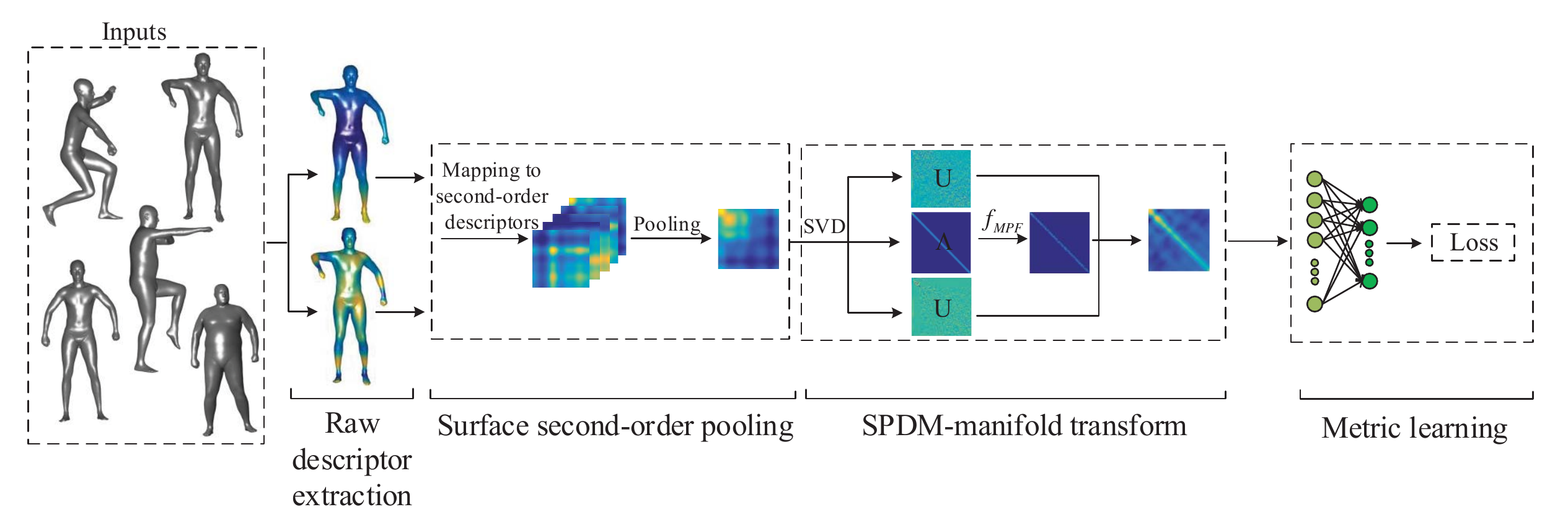}
\caption{Architecture of our proposed ST-Net. It consists of four stages, i.e., raw descriptor extraction, surface second-order pooling, SPDM-manifold transform and metric learning.}
\label{fig:PIPELINE}
\end{center}
\end{figure*}

In this work, we propose a novel learning-based spectral transform network on 3D surface to learn discriminative shape descriptor for non-rigid shape retrieval and classification. 
\textit{First}, we define a second-order pooling operation on 3D surface which models the second-order statistics of input raw descriptors on 3D surfaces. \textit{Second}, considering that the pooled second-order descriptors lie on a manifold of symmetric positive definite matrices (SPDM-manifold), we define a novel manifold transform for feature learning by learning a \emph{mixture of power function} on the singular values of the SPDM descriptors. Third, by concatenating the stages of raw descriptor extraction, surface second-order pooling, transform on SPDM-manifold and metric learning, we propose a novel network architecture, dubbed as \emph{spectral transform network} as shown in Fig.~\ref{fig:PIPELINE}, which can learn discriminative shape descriptors for non-rigid shape analysis.

To the best of our knowledge, this is the first paper that learns second-order pooling-based shape descriptors on 3D surfaces using a network architecture. Our network structure is simple and easily to be trained, and is justified to be able to significantly improve the  discriminative ability of input raw descriptors. It is adaptive to various non-rigid shapes such as watertight meshes, partial meshes and point cloud data. It achieved competitive results on challenging benchmarks for non-rigid shape retrieval and classification, e.g., $100\%$ accuracy  on SHREC'14~\cite{SHREC} dataset and the state-of-the-art accuracy on the ``range'' subset of SHREC'17~\cite{Masoumi2017SHREC} in metric of NN~\cite{Philip}.

\section{Related Works}
\subsection{Learning approach for 3D shapes.}
Deep learning is a powerful tool in computer vision, speech recognition, natural language processing, etc. Recently, it has also been extended to 3D shape analysis and achieves impressive progresses. One way for the extension is to  represent the shapes as volume data~\cite{Zhirong,HaoSu1} or multi-view data~\cite{xiangbai1} and then send them to deep neural networks. The voxel and multi-view based shape representation have been successful in rigid shape representation~\cite{HaoSu1,su2015multiview} relying on a large training dataset. Due to the operations of voxelization and 3D to 2D projection, it may lose shape details especially for non-rigid shapes with large deformations, e.g., human bodies with different poses.  An alternative way is to define the networks directly on 3D surface based on spectral descriptors as in~\cite{YiFang1,YiFang3,MMB4,masci2015geodesic}. These models  benefit from the intrinsic properties of  spectral descriptors, and utilize surface convolution or deep nueral networks, e.g., LSTM, auto-encoder,  to further learn  discriminative shape descriptors.  PointNet~\cite{qsmgpdlps3dcs17,qi2017pointnetplusplus}  is another interesting deep learning approach that directly build network using the point cloud representation of 3D shapes, which can also handle the non-rigid 3D shape classification using non-Euclidean metric.  Compared with them, we build  a novel network architecture on 3D surface from the perspectives of  second-order descriptor pooling and spectrum transform on the pooled descriptors. It is justified to be able to effectively learn surface descriptors  on SPDM-manifold.

\subsection{Second-order pooling of shape descriptors.}  
Second-order pooling operation was firstly proposed in~\cite{Carreira} showing outstanding performance in 2D vision tasks such as recognition~\cite{Catalin1} and segmentation~\cite{Carreira}. The pooled descriptors lie on a Riemannian SPDM-manifold. Due to  non-Euclidean structure of this manifold, many traditional machine learning methods based on Euclidean metrics can not be used directly. As discussed in~\cite{Vincent,Xavier}, two popular metrics on SPDM-manifold are affine-invariant metric and log-Euclidean metric. Considering the complexity, the log-Euclidean metric and its variants \cite{Catalin2,huang2016a} that embed data into Euclidean space are more widely used~\cite{Carreira,Hedi}.
The power-Euclidean transform~\cite{dryden2009non-euclidean} has achieved impressive results which theoretically approximates the log-Euclidean metric when its power index approaches zero.
The most related shape descriptors to ours for 3D shape analysis are the covariance-based descriptors~\cite{covariance1,Hedi}. In~\cite{covariance1}, they encoded the point descriptors such as angular and texture within a 3D point neighbourhood by a covariance matrix. In~\cite{Hedi}, the covariance descriptors were further incorporated into the Bag-of-Words model to represent shapes for retrieval and correspondence. In our work, we present a formal definition of second-order pooling of shape descriptors on 3D surface, and define a learning-based spectral transform on SPDM-manifold, which can effectively boost the  performance of the pooled descriptors for 3D non-rigid shape analysis.

In the following sections, we first introduce our proposed spectral transform network in Sect.~3. Then, in Sect.~4, we experimentally justify the effectiveness of the proposed network on benchmark datasets for non-rigid shape retrieval and classification. We finally conclude this work in Sect.~5.

\section{Spectral Transform Network on 3D Shapes }
We aim to learn discriminative shape descriptors for 3D shape analysis by designing a \emph{spectral transform network (ST-Net)} on 3D surface.  As illustrated in Fig.~\ref{fig:PIPELINE}, our approach consists of four stages: raw descriptor extraction, surface second-order pooling, SPDM-manifold transform and metric learning. In the followings, we will give detailed descriptions of these stages.

\subsection{Raw descriptor extraction} 
Let $\mathcal{S}$ denote the surface (either mesh or point cloud) of a given shape, in this stage, we extract descriptors from $\mathcal{S}$.  For watertight surface, we select spectral descriptors, i.e., SIHKS~\cite{MMB1} and WKS~\cite{MathieuWKS} as inputs, which are intrinsic and robust to non-rigid deformations. For partial surface and point cloud, we choose local geometric descriptors such as Localized Statistical Features(LSF)~\cite{Ohkita2012Non}. All of them are dense descriptors representing multi-scale geometric features of the shape. Note that our framework is generic, and other shape descriptors can also be used such as normals and curvatures.

\noindent
\textbf{Spectral descriptors.} 
Spectral descriptors are mostly dependent on the spectral (eigenvalues and/or eigenfunctions) of the Laplace\text{-}Beltrami operator, and they are well suited for the analysis of non-rigid shapes. Popular spectral descriptors include HKS~\cite{Jian}, SIHKS~\cite{MMB1} and WKS~\cite{MathieuWKS}. Derived from heat diffusion process, HKS~\cite{Jian} reflects the amount of heat remaining at a point after certain time. SIHKS~\cite{MMB1} is derived from HKS and it is scale-invariant.  Both of them are intrinsic but lack spatial localization capability. WKS~\cite{MathieuWKS} is another intrinsic spectral descriptor stemming from Schr\"{o}dinger equation. It evaluates the probability of a quantum particle on a shape to be located at a point under a certain energy distribution, and it is better for spatial localization. 

\noindent
\textbf{Local geometric descriptors.} 
Another kind of raw shape descriptor is local geometric descriptor, which encodes the local geometric and spatial information of the shape. We select LSF~\cite{Ohkita2012Non} as input for partial and point cloud non-rigid shape analysis, and it encodes the relative positions and angles  locally on the shape. Assuming the selected point is $s_1$, its position and normal vector are $\mathbf{p_1}$ and $\mathbf{n_1}$,  another point $s_2$ with associated position and normal vector as $\mathbf{p_2}$ and $\mathbf{n_2}$  is within the sphere of influence in a radius $r$ of $s_1$. Then a 4-tuple $(\beta_1, \beta_2, \beta_3, \beta_4)$ is computed as:
\begin{equation}
\begin{split}
&\beta_1 = \text{arctan}(\mathbf{w}\cdot{\mathbf{n_1}},\mathbf{u}\cdot{\mathbf{n_2}}),~~\beta_2 = \mathbf{v}\cdot{\mathbf{n_2}}, \\
&\beta_3 = \mathbf{u} \cdot{(\mathbf{p_2}-\mathbf{p_1})/||\mathbf{p_2}-\mathbf{p_1}||},~~\beta_4 = ||\mathbf{p_2}-\mathbf{p_1}||.
\end{split}
\end{equation}
where $\mathbf{u} = \mathbf{n_1}, \mathbf{v} = (\mathbf{p_2}-\mathbf{p_1})\times \mathbf{u}/||(\mathbf{p_2}-\mathbf{p_1})\times \mathbf{u}||, \mathbf{w} = \mathbf{u}\times \mathbf{v} $. For a local shape of $N$ points, a set of $(N-1)$ 4-tuples are computed for the center point, which are collected into a 4-dimensional joint histogram. By dividing the histogram to 5 bins for each dimension of the tuple, we have a 625-d descriptor for the center point, which encodes the local geometric information around it. 
 
Given the surface $\mathcal{S}$, we extract either spectral or geometric shape descriptors  called as raw descriptors for each point $s \in \mathcal{S}$, denoted as $\{\mathbf{h}(s)\}_{s \in \mathcal{S}}$, which are taken as the inputs of the following stage.

\subsection{Surface second-order pooling}
In this stage, we generalize the second-order average-pooling operation~\cite{Carreira} from 2D image to 3D surface, and propose a surface second-order pooling operation. Given the extracted shape descriptors $\{\mathbf{h}(s)\}_{s \in \mathcal{S}}$, the \emph{surface second-order pooling} is defined as: 
\begin{eqnarray}
H = \frac{1}{|{\mathcal{S}}|}\int_{{\mathcal{S}}}\mathbf{h}^{O_2}(s)ds,  \quad
  \mathbf{h}^{O_2}(s) = \mathbf{h}(s) \mathbf{h}(s)^\top,
\label{eqn:2p}
\end{eqnarray}
where $|{\mathcal{S}}|$ denotes the area of the surface, $\mathbf{h}^{O_2}(s)$ is the \emph{second-order descriptor} for a point $s$, and $H$ is a matrix of the pooled second-order descriptor on ${\mathcal{S}}$, which is taken as the output of this stage.

For the surface represented by discretized irregular triangular mesh, the integral operation in Eq.~(\ref{eqn:2p}) can be descretized considering the Voronoi area around each point:
\begin{equation}
H = \sum_{s \in{\mathcal{S}}}\pi(s)\mathbf{h}(s) \mathbf{h}(s)^\top, \pi(s) = \frac{a(s)}{\sum_{p\in{\mathcal{S}}}a(p)},
\label{eqn:2p0}
\end{equation}
where $s$ denotes a discretized point on ${\mathcal{S}}$ with its Voronoi area as $a(s)$. In our work, we compute $a(s)$ as in~\cite{Mathieu}.
For the shapes composed of point cloud, Eq.~(\ref{eqn:2p}) can be descretized as average pooling of the second-order information:
\begin{equation}
H = \frac{1}{|\mathcal{S} |} \sum_{s \in{\mathcal{S}}}\mathbf{h}(s) \mathbf{h}(s)^\top,
\label{eqn:2pd}
\end{equation}
where $|\mathcal{S}|$ denotes the number of points on the surface.

The pooled second-order descriptors represent 2nd-order statistics of raw descriptors over the 3D surfaces.  It is obvious that $H$ is a symmetric positive definite matrix (SPDM), which lies on a non-Euclidean manifold of SPDM.

\subsection{SPDM-manifold transform}
This stage, i.e., SPDM-T stage, performs non-linear transform on the singular values of the pooled second-order descriptors, and it is in fact a spectral transform on the SPDM-manifold. This transform will be discriminatively learned for specific task enforced by the loss in the next metric learning stage. 

\noindent
\textbf{Forward computation.}  Assuming that we have a symmetric positive definite matrix $H$, by singular value decomposition, we have:
\begin{equation}
H = U\Lambda U^\top.
\end{equation}
We first normalize the singular values of $H$, i.e., the diagonal values of $\Lambda$, by $\mathcal{L}_2$-normalization, achieving $\{\tilde{\Lambda}_{ii}\}_{i=1}^{N_{\Lambda}}$, where $N_{\Lambda}$ is the number of singular values, then perform non-linear transform on $\{\tilde{\Lambda}_{ii}\}_{i=1}^{N_{\Lambda}}$. Inspired by polynomial function, we propose the following transform:
\begin{equation}
\Lambda'= \mbox{diag} \{f_{MPF}(\tilde{\Lambda}_{11}), \cdots, f_{MPF}(\tilde{\Lambda}_{N_{\Lambda}N_{\Lambda}})\},
\end{equation}
where $\mbox{diag}\{\cdot\}$ is a diagonal matrix with input elements as its diagonal values, $f_{MPF}(\cdot)$ is a \emph{mixture of power function}:
\begin{equation}
f_{MPF}(x)= \sum_{i=0}^{N_m}\gamma_{i} x^{\alpha_i},~\alpha_i\in [0,1],
\end{equation}
where  $\{\alpha_i\}_{i=0}^{N_m}$ are $N_m+1$ samples with uniform intervals  in range of $[0,1]$, $\Gamma = (\gamma_0,\gamma_1,\cdots,\gamma_{N_m})^{\top}$ is a vector of combination coefficients and required to satisfy:
\begin{equation}
\Gamma^\top \mathbf{1}  = 1,~\Gamma\geq \mathbf{0}.
\label{eqn:norm}
\end{equation} 
To meet these requirements, the coefficients are defined as:
\begin{equation}
\gamma_i = \frac{e^{\omega_i}}{\sum_{j=0}^{N_m}e^{\omega_j}},~i=0,1,\cdots,N_m.
\end{equation}
Then we instead learn the parameters in $\Omega=(\omega_0,\omega_1,\cdots,\omega_{N_m})^{\top}$ to determine $\Gamma$.

After this transform, a new singular value matrix $\Lambda'$ is derived. Combining it with the original singular vector matrix $U$, we get the transformed descriptor $H'$ as:
\begin{equation}
\begin{split}
H' = U \Lambda' U^{\top} =  U \mbox{diag} \{f_{MPF}(\tilde{\Lambda}_{11}), \cdots, f_{MPF}(\tilde{\Lambda}_{N_{\Lambda}N_{\Lambda}})\} U^{\top}.
\end{split}
\end{equation}
$H'$ is also a symmetric positive definite matrix. Due to the symmetry of $H'$, the elements of its upper triangular $g(H')$ are kept as the output of this stage, where $g(\cdot)$ is an operator vectorizing the upper triangular elements of a matrix. 

\noindent
\textbf{Backward propagation.} As proposed in~\cite{Catalin2}, matrix back-propagation can be performed for SVD decomposition. Let $(\cdot)_{diag}$ denote an operator on matrix that sets all non-diagonal elements to $0$, $(\cdot)_{Gdiag}$ be an operator of vectorizing the diagonal elements of a matrix, $g^{-1}(\cdot)$ be the inverse operator of $g(\cdot)$, $\odot$ be the Hadamard product operator.
For backward propagation, assuming the partial derivative of loss $L$ with respect to $g(H')$ as $\frac{\partial L}{\partial g(H')}$,  we have:
\begin{equation}
\frac{\partial L}{\partial \Lambda'} = (U^{\top} g^{-1}(\frac{\partial L}{\partial g(H')}) U)_{diag},
\label{equ:d1}
\end{equation}
\begin{equation}
(\frac{\partial L}{\partial \Gamma})_i = (\tilde{\Lambda}^{\alpha_i}\frac{\partial L}{\partial \Lambda'})_{Gdiag}^{\top} \mathbf{1},~i=0,1,...,N_m,
\label{equ:d2}
\end{equation}
\begin{equation}
\frac{\partial L}{\partial \Omega} = (\frac{\partial L}{\partial \Gamma} -\Gamma^{\top} \frac{\partial L}{\partial \Gamma} )\odot\Gamma.
\label{equ:d3}
\end{equation}
The partial derivative of loss function $L$ with respect to the parameter $\Omega$ can be derived by successively computing Eqs.~(\ref{equ:d1}), (\ref{equ:d2}), (\ref{equ:d3}). Please refer to supplementary material for gradient computations.

\noindent
\textbf{Analysis of SPDM-T stage.} The pooled second-order descriptor $H$ lies on the SPDM-manifold, and the popular transform on this manifold is log-Euclidean transform~\cite{dryden2009non-euclidean}, i.e., $H' = \mbox{log}(H)$. However, it is unstable when the singular values of $H$ are near or equal to zero.  The logarithm-based transforms such as $H' = \mbox{log}(H+\epsilon I)$~\cite{Catalin2} and $H' = \mbox{log}(\mbox{max}\{H, \epsilon I\})$~\cite{huang2016a} are proposed to overcome this unstability, but they need a positive constant regularizer $\epsilon$  which is difficult to set. 
The power-Euclidean metric~\cite{dryden2009non-euclidean} theoretically approximates the log-Euclidean metric when its power index approaches zero while being more stable. Our proposed mixture of power function $f_{MPF}(\cdot)$ is an extension of power-Euclidean transform that takes it as a special case. The SPDM-T stage learns an effective transform in the space spanned by the power functions adaptively using a data-driven approach. Furthermore, the mixture of power function $f_{MPF}(\cdot)$ is constrained to be nonlinear and retains non-negativeness and order of the eigenvalues (i.e., singular values of a symmetric matrix).

From a statistical perspective,  $H$ can be seen as a covariance matrix of input descriptors on 3D surface. Geometrically, its eigenvectors in columns of $U$ construct a coordinate system, its eigenvalues reflect feature variances projected to eigenvectors.  By transforming these projected variances (eigenvalues),  $f_{MPF}(\cdot)$ implicitly tunes the statistics distribution of input raw descriptors in pooling region when training. Since the entropy of Gaussian distribution with covariance $H \in R^{d \times d}$ is $\mathcal{E}(H) = \frac{1}{2}(d + \log(2\pi) + \log\prod_{i}{\Lambda}_{ii})$, transforming eigenvalues ${\Lambda}_{ii}$ by $f_{MPF}(\cdot)$  implicitly tune the entropy of distribution of raw descriptors on 3D surface.

\subsection{Metric learning}

With the transformed descriptors $g(H')$ as input, we embed them into a low-dimensional space where the descriptors are well grouped or separated with the guidance of labels. To prove the effectiveness of the SPDM-T stage, we design a shallow neural network to achieve the metric learning stage. We first normalize the input $g(H')$  by $\mathcal{L}_2$-normalization, achieving $\tilde{g}(H')$, then add a fully connected layer:
\begin{equation}
F = W \tilde{g}(H'),
\end{equation}
where $W$ is a matrix in size of $D_m \times D_p$. $F$ is the descriptor of the whole shape. We further send $F$ into loss function for specific shape analysis task to enforce the discriminative ability of shape descriptor.  In this work, we focus on shape retrieval and classification. 
We next discuss the loss functions. 

\noindent
\textbf{Shape retrieval.} Given a training set of shapes, the loss for shape retrieval is defined on all the possible triplets of shape descriptors $\mathcal{T}_R=\{F_i, F_i^{Pos}, F_i^{Neg}\}$, where $i$ is the index of  triplet, $F_i^{Pos}$ and $F_i^{Neg}$ are two shape descriptors with same and different labels w.r.t. the target shape descriptor $F_i$ respectively:
\begin{equation}
\begin{split}
L = \sum_{i=1}^{|\mathcal{T}_R|}~&(\mu + ||F_i - F_i^{Pos}|| - ||F_i - F_i^{Neg}||)_{+}^2
+\eta||F_i - F_i^{Pos}||,
\end{split}
\end{equation}
where $|\mathcal{T}_R|$ is the number of triplets in $\mathcal{T}_R$, $||\cdot||$ is $\mathcal{L}_2$-norm, $\mu$ is the margin, $(\cdot)_+ = \max\{\cdot,0\}$ and $\eta$ is a constant to balance these two terms.

\noindent
\textbf{Shape classification.} We construct the cross-entropy loss for shape classification. Given the learned descriptor $\{F_i\}$ with their corresponding labels as $\{y_i\}$, we first add a fully connected layer after ${F_i}$ to map the features to scores for different categories, and then followed by a softmax layer to predict the probability of a shape belonging to different categories, and the probabilities of all training shapes are denoted as $\mathcal{T}_C=\{\widetilde{F}_i\}$. The loss function is defined as:
\begin{equation}
\begin{split}
L = \sum_{i=1}^{|\mathcal{T}_C|}~\sum_{j=1}^{M}  y_i^j \text{log}(\widetilde{F}_i^j) + (1-y_i^j)\text{log}(1-\widetilde{F}_i^j)
\end{split}
\end{equation}
where $|\mathcal{T}_C|$ is the number of training shapes, $i$ and $j$ indicate the shape and category respectively, and $M$ is the total number of  categories.

The combination of  fully connected layer and loss function results in  a metric learning problem. Minimizing the loss function embeds the shape descriptors into a lower-dimensional space, in which the learned shape descriptors are enforced to be discriminative for specific shape analysis task.

\subsection{Network training}
For the task of shape retrieval, each triplet of shapes $\{F_i, F_i^{Pos}, F_i^{Neg}\}$ is taken as a training sample and multiple triplets are taken as a batch for training with mini-batch stochastic gradient descent optimizer~(SGD). 
For shape classification, the network is also trained by mini-batch SGD. 
To train the network, the raw descriptor extraction and second-order pooling stage as well as the SVD decomposition can be computed off-line, and the learnable parameters in ST-Net are $\Omega$ in SPDM-T stage and $W$ in metric learning stage and the later fully connected layer (for classification). 
For the non-linear transform in  SPDM-T stage, we set ${N_m}=10$, $\alpha_i = \frac{i}{10},~i \in \{0,1,...,10\}$. The gradients of loss are back-propagated to the SPDM-T stage.

\section{Experiments}
In this section, we evaluate the effectiveness of our ST-Net, especially the surface second-order pooling and SPDM-T stages, for 3D non-rigid shape retrieval and classification. We test our model on watertight and partial mesh datasets as well as point cloud dataset. We will successively introduce the datasets, evaluation methodologies, quantitative results and the evaluation of our SPDM-T stage.

\subsection{Datasets and evaluation methodologies}
Considering that our network is designed for non-rigid shape analysis, we evaluate it for shape retrieval on SHREC'14~\cite{SHREC} and SHREC'17~\cite{Masoumi2017SHREC} datasets, and we test our architecture for shape classification on SHREC'15~\cite{Lian2015SHREC} dataset. All of them  are composed of non-rigid shapes with various deformations.

\noindent
\textbf{SHREC'14.}  This dataset includes two datasets of Real and Synthetic human data respectively, both of which are composed of watertight meshes. The Real dataset comprises of $400$ meshes from $40$ human subjects in $10$ different poses. The Synthetic dataset consists of 15 human subjects with $20$ poses, resulting in a dataset of 300 shapes.  We will try three following experimental settings, which will be refered as ``setting-$i$" ($i =1, 2, 3$). In setting-$1$, $40\%$ and $60\%$ shapes are used for training and test respectively as in~\cite{Ioannis}. In setting-$2$, an independent training set\footnote{\url{http://www.cs.cf.ac.uk/shaperetrieval/download.php}} including unseen shape categories is taken as training set and the Real dataset of SHREC'14 is used for test. In setting-3,  $30\%$ of the classes are randomly selected as the training classes and remaining $70\%$ classes are used for test as in  Litman~\cite{SHREC}. Both setting-$2$ and $3$ are challenging because the shapes in training and test sets are disjoint  in shape categories.

\noindent
\textbf{SHREC'15.} This dataset includes 1200 watertight shapes of 50 categories, each of which contains 24 shapes with various poses and topological structures. To compare with the state-of-the-art PointNet++~\cite{qi2017pointnetplusplus} on the dataset, we use the experimental setting in \cite{qi2017pointnetplusplus}, i.e., treating the shapes as point cloud data, and using 5-fold cross-validation to test the accuracy for shape classification. 

\noindent
\textbf{SHREC'17.} This dataset is composed of two subsets, i.e., ``holes'' and ``range'', which contain meshes with holes and range data respectively.  
We use the provided standard splits for training / test. The ``holes'' subset consists of 1216 training and 1078 test shapes, and the ``range'' subset consists of 1082 training and 882 test shapes.

\noindent
\textbf{Evaluation methodologies.} For  non-rigid shape retrieval, we evaluate results by NN (Nearest Neighbor), 1-T (First-Tier), 2-T (Second-Tier), and DCG (Discounted Cumulative Gain)~\cite{Philip}. For non-rigid shape classification, the results are evaluated by classification accuracy, i.e., the percentage of correctly classified shapes.

\subsection{Results for non-rigid shape retrieval on watertight dataset}
\begin{table*}[t]
\caption{Retrieval results on SHREC'14 Synthetic and Real dataset in setting-1 (in $\%$). We show the results of ST-Net as well as the baselines and state-of-the-art CSDLMNN~\cite{Ioannis}.}
\newcommand{\tabincell}[2]{\begin{tabular}{@{}#1@{}}#2\end{tabular}}
\begin{center}
\begin{tabular}{|l||  c c   c    c|| c   c  c c|}
\hline
\multirow{2}{*}{Method}              &\multicolumn{4}{c||}{Synthetic}                                &\multicolumn{4}{c|}{Real} \\
\cline{2-9}                         & NN        & 1-T          &  2-T          & DCG        & NN     & 1-T      &  2-T           & DCG\\
\hline
CSDLMNN~\cite{Ioannis}&99.7        &98.0         & 99.9      & 99.6       &97.9    &  92.8   &  98.7       &  97.6\\
Surf-O$_1$ &82.7  & 77.1 & 84.3 & 83.6     &54.2  & 52.6 & 57.9 & 55.0  \\   
Surf-O$_2$ &87.3  & 84.2 & 89.2 & 87.8     &61.1  & 57.2 & 64.0 &  63.2  \\   
Surf-O$_1$-ML &\textbf{100} & 96.9        &  99.9       & 99.7      &96.7   &  91.9   &  98.3       &  96.9\\
Surf-O$_2$-ML &\textbf{100} &\textbf{100}  &\textbf{100}   &\textbf{100} &98.8   &  96.1   &  99.6       &  99.9\\
ST-Net        &\textbf{100} &\textbf{100}  &\textbf{100}  &\textbf{100} &\textbf{100} &\textbf{99.8}  &\textbf{100} &\textbf{99.9}\\
\hline
\end{tabular}
\end{center}
\label{tab:tab1}
\end{table*}
For non-rigid shape retrieval on SHREC'14 datasets, we select SIHKS and WKS as input raw descriptors. We discretize the Laplace-Beltrami operator as in~\cite{Mathieu}, and compute $50$-d SIHKS and $100$-d WKS. In the surface second-order pooling stage, the descriptors are computed by  Eq.~(\ref{eqn:2p0}). When training the ST-Net for shape retrieval, the batch size, learning rate and margin $\mu$ are set as $5$, $20$ and $60$ respectively. The descriptor of every shape is $100$-d, i.e., $D_m = 100$.  In the loss function,  $\eta$ is set as $1$. 

To justify the effectiveness of our architecture, we compare the following different variants of descriptors for shape retrieval. (1) \emph{Surf-O$_1$}:  pooled raw descriptors on surfaces. (2) \emph{Surf-O$_2$}:  pooled  second-order descriptors on surfaces. (3) \emph{Surf-O$_1$-ML}:  descriptors of Surf-O$_1$ followed by a metric learning stage. (4)  \emph{Surf-O$_2$-ML}: descriptors of Surf-O$_2$ followed by a metric learning stage.
For retrieval task, the descriptors of \emph{Surf-O$_1$} and \emph{Surf-O$_2$} are directly used for retrieval based on Euclidean distance. 
In Table~\ref{tab:tab1}, we report the results in experimental setting-1 of these descriptors as well as state-of-the-art CSDLMNN~\cite{Ioannis} method. As shown in the table, the increased accuracies from Surf-O$_1$ to Surf-O$_2$, and that from  Surf-O$_1$-ML to Surf-O$_2$-ML indicate the effectiveness of the surface second-order pooling stage. The improvements from Surf-O$_2$-ML to ST-Net demonstrate the advantage of the SPDM-T stage.  Our full ST-Net achieves 100\% accuracy in NN (i.e., the percentage of retrieved nearest neighbor shapes belonging to the same class as queries) on SHREC'14 Synthetic and Real datasets. Compared with state-of-the-art CSDLMNN~\cite{Ioannis} method, the competitive accuracies justify the effectiveness of our method. 

Table~\ref{tab:tab3} presents results in mean average precision~(mAP) on SHREC'14 Real and Synthetic datasets in setting-1 compared with RMVM~\cite{RMVM}, CSDLMNN~\cite{Ioannis}, in which CSDLMNN is a state-of-the-art approach for this task. For our proposed ST-Net, we randomly split the training and test subsets five times and report the average mAP with standard deviations shown in brackets. In the table, we also present the baseline results of our ST-Net to justify the effectiveness of our architecture. Our ST-Net achieves highest mAP on both datasets, demonstrating its effectiveness for watertight non-rigid shape analysis.
In Table~\ref{tab:tabset2}, we also show the results on  SHREC'14 Real dataset using setting-2  and  setting-3, which are more challenging since the training and test sets have disjoint shape categories.  ST-Net still significantly outperforms the baselines of Surf-O$_1$-ML and Surf-O$_2$-ML, and achieves high accuracies. Our ST-Net significantly outperforms Litman~\cite{SHREC} using the experimental setting-3.

\begin{table}
\caption{Results for shape retrieval in setting-1 evaluated by mAP in $\%$ on SHREC'14 Synthetic and Real datasets. Our ST-Net performs best on both datasets.}
\begin{center}
\begin{tabular}{|l|| c| c|  c| c| c| c| c|}
\hline
Dataset & RMVM\cite{RMVM} & CSDLMNN\cite{Ioannis}  & Surf-O$_1$ & Surf-O$_2$ & Surf-O$_1$-ML & Surf-O$_2$-ML  & ST-Net\\
\hline
Synthetic & 96.3 & 99.7  & 82.7 &85.3 &93.6 & 97.1  & \textbf{100}(0)\\
Real      & 79.5 & 97.9  & 50.8 &59.4 &90.5 & 95.3  & \textbf{99.9}(0.1)\\
\hline
\end{tabular}
\end{center}
\label{tab:tab3}
\end{table}
\begin{table}
\begin{center}
\caption{Results for shape retrieval on SHREC'14 Real dataset in setting-2 and -3 (in $\%$). ST-Net achieves the best performance.}
\begin{tabular}{|l||c|c|c|c|c|c|}
\hline
         &method & NN & 1-T & 2-T & EM & DCG \\
          \hline
\multirow{3}{*}{Setting-2} & Surf-O$_1$-ML & 45.75 & 35.25& 59.08& 34.76& 63.84\\
 &Surf-O$_2$-ML& 80.25 & 63.94 & 78.39 & 40.94 & 80.21 \\
 &ST-Net& \textbf{85.75} & \textbf{71.33} & \textbf{88.92} & \textbf{43.20} & \textbf{88.29} \\
\hline
\multirow{4}{*}{Setting-3} & Litman\cite{SHREC} & 79.3 & 72.7 & 91.4 & 43.2 & 89.1 \\
&Surf-O$_1$-ML& 54.29  & 46.67 & 70.91  & 37.60  & 71.71  \\
&Surf-O$_2$-ML& 88.76  & 82.01  & 96.47  & 42.82  & 90.75  \\
&ST-Net& \textbf{92.53}(1.49) & \textbf{84.78}(2.43) & \textbf{96.93}(1.17) & \textbf{43.86}(0.36) & \textbf{93.85}(1.63) \\
\hline
\end{tabular}
\label{tab:tabset2}
\end{center}
\end{table}

\begin{figure} [t]
\begin{center}
\includegraphics[width=0.8\linewidth,height=2.7in]{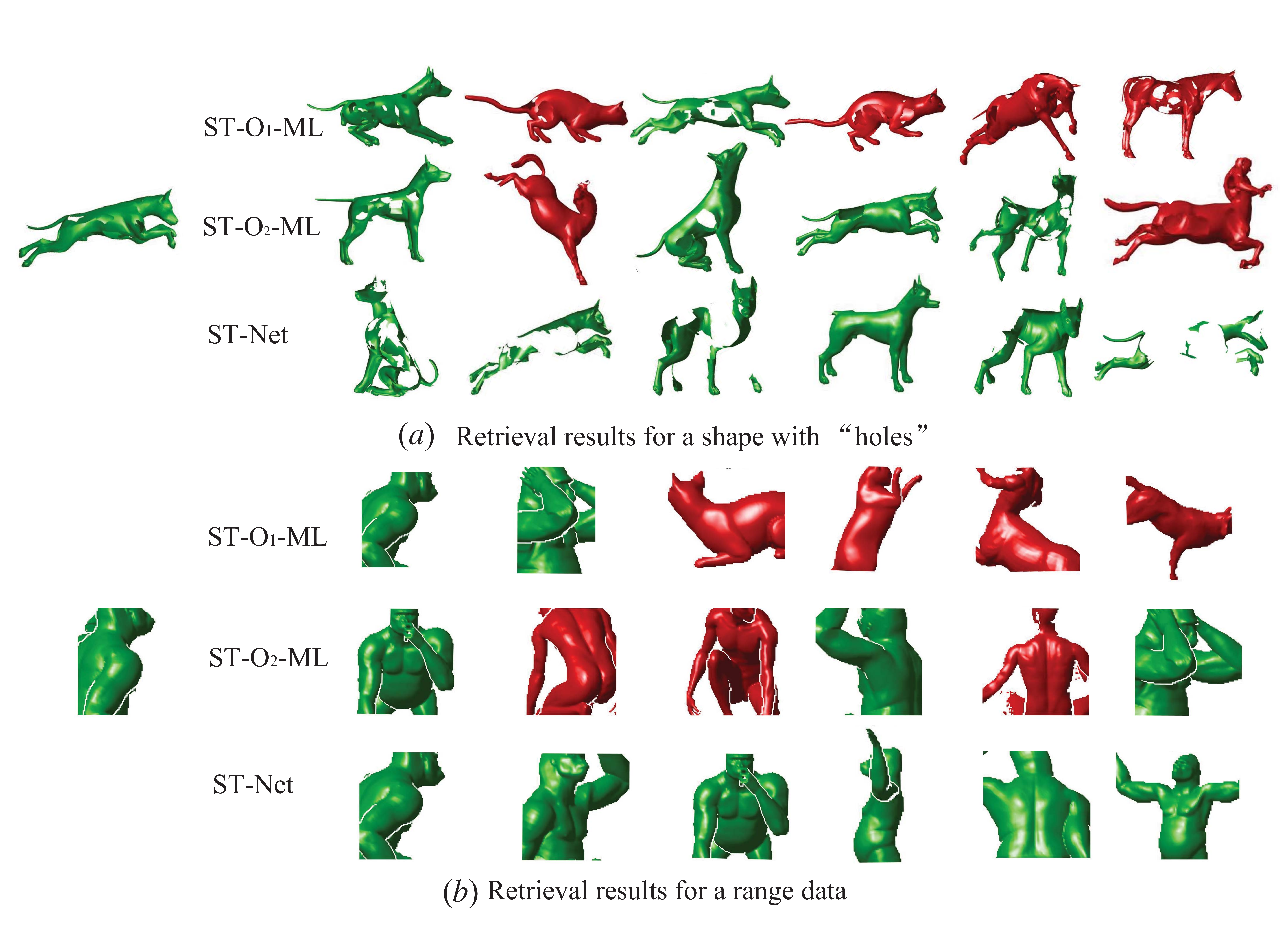}
\caption{Instances of retrieved top ranked shapes with interval of 5 in ranking index.  The wrongly retrieved shapes are shown in  red.}
\label{fig:retreival_results}
\end{center}
\end{figure}

\subsection{Results for  non-rigid shape retrieval on partial dataset}
We now evaluate our approach on non-rigid partial shapes in subsets of ``holes'' and ``range'' of SHREC'17 dataset. Considering that the shapes are not watertight surface, we use local geometric descriptors as raw descriptors. We select 3000 points uniformly as in~\cite{Osada2002Shape} for every shape and compute 625-d LSF as inputs. In the surface second-order pooling stage, the descriptors are computed by Eq.~(\ref{eqn:2pd}). When training the ST-Net for shape retrieval, the batch size, learning rate and margin $\mu$ are set as $5$, $100$ and $60$ respectively. The descriptor of every shape is $300$-d, i.e., $D_m = 300$.  In the loss function, the constant $\eta$ is set as $1$. 

In Table~\ref{tab:tab55}, we first compare our ST-Net with the baselines, i.e., Surf-O$_1$, Surf-O$_2$, Surf-O$_1$-ML and Surf-O$_2$-ML to show the effectiveness of our network architecture.  The raw LSF descriptors after second-order pooling without learning, i.e., Surf-O$_2$, produces the accuracy of 69.0\% and 71.7\% in NN on ``holes'' and ``range'' subsets respectively. With the metric learning stage, the results are increased to 75.9\% and 77.6\% for these two subsets. The ST-Net with both SPDM-T stage and metric learning stage increases the accuracies to be 96.1\% and 97.3\% respectively, with around 20 percent improvement than the results of Surf-O$_2$-ML. This clearly shows the effectiveness of our defined SPDM-T stage for enhancing the discriminative ability of shape descriptors. 
Moreover, the performance increases from Surf-O$_1$ to Surf-O$_2$ and from Surf-O$_1$-ML to Surf-O$_2$-ML justify the effectiveness of surface second-order pooling stage over the traditional first-order pooling (i.e., average pooling) on surfaces.

In Table~\ref{tab:tab55}, we also compare our results with the methods that participate to the SHREC'17 track, i.e., DLSF~\cite{Furuya2016Deep}, 2VDI-CNN, SBoF~\cite{Litman2014Supervised}, and BoW+RoPS, BoW+HKS,  RMVM\cite{RMVM}, DNA~\cite{reuter2006laplacebeltrami}, etc., and  the results of their methods are from~\cite{Masoumi2017SHREC}. In DLSF~\cite{Furuya2016Deep}, they design their deep network by first training a E-block and then training a AC-block. The method of 2VDI-CNN is based on multi-view projections of 3D shapes and deep GoogLeNet\cite{szegedy2015going}  for shape descriptor learning. The method of SBoF~\cite{Litman2014Supervised} trains a bag-of-features model using sparse coding. The methods of BoW+RoPS, BoW+HKS combine BoW model with shape descriptors of RoPS and HKS  for shape retrieval. 
As shown in the table, our ST-Net ranks  second on the ``holes'' subset and achieves  the highest accuracy in NN, 2-T and DCG on the ``range'' subset, demonstrating its effectiveness for partial non-rigid shape analysis.  Compared with the deep learning-based methods of DLSF~\cite{Furuya2016Deep}, 2VDI-CNN, our network architecture is shallow and simple to implement but achieves competitive performance. 

In Figure~\ref{fig:retreival_results}, we show the top retrieved shapes with interval of 5 in ranking index given query shapes on the leftmost column,  and the examples in sub-figures  (a) and (b) are respectively from  SHRES'17 ``holes'' and ``range'' subsets.  These shapes are with large non-rigid deformations. The examples show that ST-Net enables to effectively retrieve the correct shapes even when the shapes are range data or  with large holes.

\begin{table}
\begin{center}
\caption{Shape retrieval results on SHREC'17 non-rigid datasets. The deep learning-based methods are signed with ``*''. The values are formatted as percentage.}
\newcommand{\tabincell}[2]{\begin{tabular}{@{}#1@{}}#2\end{tabular}}
\centering
\subtable[The ``holes'' subset]{
\begin{tabular}{|l||  c  c  c   c|  }
\hline
Meshod &NN & 1-T & 2-T & DCG \\
\hline
DLSF*~\cite{Furuya2016Deep}  & \textbf{100} & \textbf{97.1} &\textbf{99.9} &\textbf{99.8}\\ 
2VDI-CNN*     &90.6  &81.8 &93.7  &95.4 \\
SBoF~\cite{Litman2014Supervised}         &81.5  &32.6 &49.4  &78.0 \\
BoW+HKS      &57.8  &26.1 &43.6  &72.5 \\
BoW+RoPS   &60.7  &91.8 &97.0  &96.8 \\
RMVM\cite{RMVM} &39.2 &22.6 &40.2 &67.9\\
DNA~\cite{reuter2006laplacebeltrami}  &7.8  &16.3 & 34.8  &63.2\\ 
\hline
Surf-O$_1$  & 66.8&23.6 &37.5 &71.3 \\
Surf-O$_2$  & 69.0&24.0 &38.8 &71.7 \\
Surf-O$_1$-ML  & 73.2&29.4 &46.7 & 75.4 \\
Surf-O$_2$-ML&75.9 &49.4 & 72.3& 83.2  \\
ST-Net & 96.1&85.8 &95.7 &97.7 \\
\hline
\end{tabular}
}
\qquad
\subtable[The ``range'' subset]{        
\begin{tabular}{|l||  c  c  c   c|}
\hline
Meshod & NN & 1-T & 2-T & DCG\\
\hline
2VDI-CNN*    &96.9 &90.6 &97.7 &98.0\\
SBoF~\cite{Litman2014Supervised}         &81.1 &31.7 &51.0 &76.9\\
BoW+RoPS    &51.5 &\textbf{91.5} &95.9 &96.0\\
BoW+HKS       &51.9 &32.6 &53.7 &73.6 \\
DNA~\cite{reuter2006laplacebeltrami} &13.0 &18.3 &36.6 &64.0\\
\hline
Surf-O$_1$  &69.8 &26.1& 39.4 &71.3 \\
Surf-O$_2$  &71.7 &25.7 & 40.2&71.3 \\
Surf-O$_1$-ML  & 75.6&27.8 &47.7 &73.8 \\
Surf-O$_2$-ML &77.6 &58.2 &79.7 &87.2 \\
ST-Net  &\textbf{97.3} &86.2 &\textbf{97.9} &\textbf{98.4} \\
\hline
\end{tabular}
}
\label{tab:tab55}
\end{center}
\end{table}

\subsection{Results for  non-rigid shape classification on point cloud data}
In this section, we mainly aim to compare our approach with state-of-the-art deep network of PointNet++~\cite{qi2017pointnetplusplus} for non-rigid shape classification by point cloud representation. We compare on SHREC'15 non-rigid shape dataset for classification, and PointNet++ reported state-of-the-art results on this dataset. For every shape, we uniformly sample 3000 points as~\cite{Osada2002Shape}, and take the 625-d LSF as raw descriptors. The second-order descriptors are pooled by Eq.~(\ref{eqn:2pd}).  When training the ST-Net, the batch size, learning rate are set as $15$ and $1$.

We compare ST-Net with baseline architectures of Surf-O$_1$, Surf-O$_2$, Surf-O$_1$-ML, Surf-O$_2$-ML in Table~\ref{tab:tab6}. For the ST-Net, we perform 5-fold cross-validation for 6 times (each time using a different  train/test split), and the average accuracy is reported with standard deviation shown in bracket. For classification task, the descriptors are sent to classification loss (see Sect.~3.4) for classifier training. The raw descriptors using average pooling, i.e., Surf-O$_1$, achieves 85.58\%  in classification accuracy. Our ST-Net  achieves $97.37\%$ accuracy, which shows the effectiveness of our network architecture.

In Table~\ref{tab:tab6}, we also present the classification accuracies of  deep learning-based methods of DeepGM~\cite{Luciano2017Deep} and PointNet++~\cite{qi2017pointnetplusplus}. DeepGM~\cite{Luciano2017Deep} learns deep features from geodesic moments by stacked autoencoder. PointNet++~\cite{qi2017pointnetplusplus} is pioneering work for deep learning on point clouds based on a well designed PointNet~\cite{qsmgpdlps3dcs17} recursively on a nested partition of the input point set. 
Compared with the state-of-the-art method of PointNet++~\cite{qi2017pointnetplusplus}, the classification accuracy of our ST-Net is higher. 

\begin{table}[t]
\caption{Results for shape classification (in $\%$) on SHREC'15. For ST-Net, we repeat 5-fold cross-validation for 6 times and report the average accuracy with standard deviation in bracket.}
\newcommand{\tabincell}[2]{\begin{tabular}{@{}#1@{}}#2\end{tabular}}
\begin{center}
\begin{tabular}{|l|| c| c|  c| c| c| c| c|}
\hline
  & DeepGM~\cite{Luciano2017Deep} & PointNet++~\cite{qi2017pointnetplusplus} & Surf-O$_1$ & Surf-O$_2$ & Surf-O$_1$-ML & Surf-O$_2$-ML & ST-Net\\
\hline
Acc & 93.03 & 96.09  & 85.58&89.84 & 91.00 & 93.08  & \textbf{97.37}(0.97)\\
\hline
\end{tabular}
\end{center}
\label{tab:tab6}
\end{table}
 
\subsection{Evaluation for SPDM-manifold transform}
SPDM-T is an essential stage in our ST-Net.  Besides the analysis in Sect.~3.3, we evaluate and visualize the learned SPDM-manifold transform in this subsection.

We first evaluate the effects of different transforms in the SPDM-T stage quantitatively on SHREC'14 Real dataset in setting-3.  These transforms include power-Euclidean ($1/2$-pE)~\cite{dryden2009non-euclidean}, i.e., $y = \sqrt{x}$,  and logarithm-based transforms: $y = \mbox{log}(x)$ (L-E)~\cite{dryden2009non-euclidean}, $y = \mbox{log}(x+\epsilon)$ (L-R)~\cite{Catalin2}   and $y = \mbox{log}(\mbox{max}\{x, \epsilon\})$ (L-M-R)~\cite{huang2016a}.  Besides the transforms mentioned above, we also present the results of $\mathcal{L}_2$- Normalization ($\mathcal{L}_2$-N) and Signed Squareroot +$\mathcal{L}_2$- Normalization (SSN)~\cite{Tsungyu}. These compared results in Table~\ref{tab:tab4} are produced by ST-Net with $f_{MPF}(\cdot)$ fixed as these transforms. The results are measured by NN and $1$-T. It is shown that our learned transform achieves the best results. Some of these compared transforms, such as L-E, L-R, L-M-R, perform well in the training set but worse in the test set, and our proposed transform  prevents overfitting. 

\begin{figure}
\begin{center}
\includegraphics[width=0.9\linewidth,height=1.5in]{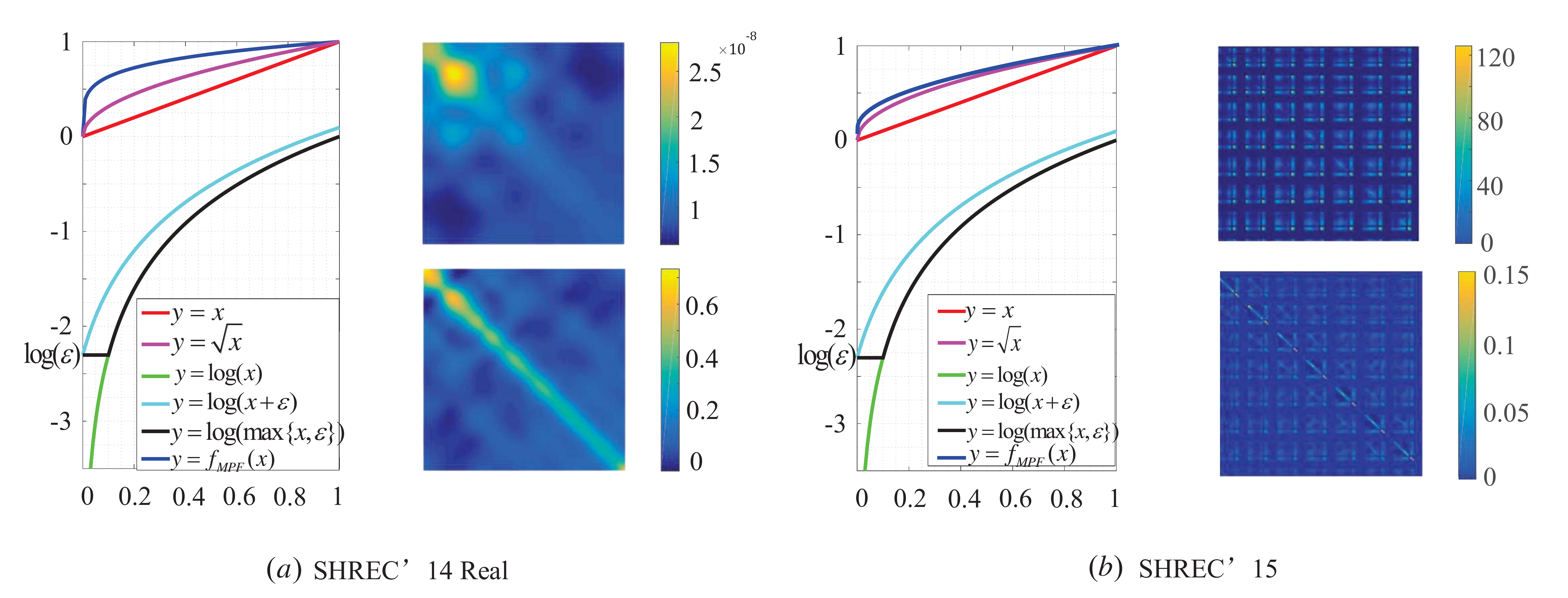}
\caption{Illustrations of learned $f_{MPF}(\cdot)$ in SPDM-T stage. Each sub-figure shows transform curves including our learned $f_{MPF}(\cdot)$ and an example of pooled second-order descriptors before (upper-right images) and after (lower-right images) transform using learned $f_{MPF}(\cdot)$.} 
\label{fig:BeforeAfterSMT}
\end{center}
\end{figure}

We then visually show the learned transform function $f_{MPF}(\cdot)$ in  Fig.~\ref{fig:BeforeAfterSMT}. We draw the curves of learned  $f_{MPF}(\cdot)$  and show examples of pooled second-order  descriptors before and after transform using  $f_{MPF}(\cdot)$ on SHREC'14 Real dataset  for retrieval (Fig.~\ref{fig:BeforeAfterSMT}(a)) and  SHREC'15 dataset  for classification (Fig.~\ref{fig:BeforeAfterSMT}(b)). 
As shown in the curves, our learned $f_{MPF}(\cdot)$ increases the eigenvalues and increases more on the smaller eigenvalues of the pooled second-order descriptors. According to the analysis in Sect. 3.3, the learned transform  $f_{MPF}(\cdot)$ increases the entropy of distribution of input raw descriptors, resulting in more discriminative shape descriptors using metric learning as shown in the experiments. 
In each sub-figure of Fig.~\ref{fig:BeforeAfterSMT}, compared with traditional fixed transforms, our net can adaptively learn transforms $f_{MPF}(\cdot)$ for different tasks by discriminative learning. We also show the pooled second-order descriptors before (upper-right images) and after (lower-right images) the transform of $f_{MPF}(\cdot)$ in the sub-figures, and the values around diagonal elements  are enhanced after transform. 

\begin{table}[t]
\caption{Results for shape retrieval on SHREC'14 Real dataset in setting-3~(in $\%$). Our transform performs the best in both of the training and test sets.}
\newcommand{\tabincell}[2]{\begin{tabular}{@{}#1@{}}#2\end{tabular}}
\begin{center}
\begin{tabular}{|l|  c||  c|  c|   c|  c|   c|    c|  c|}
\hline
\multicolumn{2}{|c||}{Metric}    &L-E~\cite{dryden2009non-euclidean}  & L-R~\cite{Catalin2}   & L-M-R~\cite{huang2016a}  & $\mathcal{L}_2$-N  & SSN~\cite{Tsungyu}   &  $1/2$-pE~\cite{dryden2009non-euclidean}  & Proposed   \\
\hline
\multirow{2}{*}{NN} & train & 100   & 93.33 & 95.00 & 95.83 & 97.50 & 99.17 & 98.33  \\
\cline{2-2} & test          & 61.07 & 65.00 & 65.36 & 82.14 & 84.64 & 88.57 & \textbf{92.50}  \\
\hline
\multirow{2}{*}{1-T} &train & 93.15 & 82.04 & 83.61 & 87.69 & 95.37 & 96.48 &  96.67 \\
\cline{2-2} & test          & 53.17 & 49.40 & 50.36 & 70.20 & 75.32 & 77.30 &  \textbf{85.20} \\
\hline
\end{tabular}
\end{center}
\label{tab:tab4}
\end{table}

\section{Conclusion}
In this paper, we proposed a novel spectral transform network for 3D shape analysis based on surface second-order pooling and spectral transform on SPDM-manifold. The network is simple and shallow. Extensive experiments on benchmark datasets show that it can significantly boost  the discriminative ability of input shape descriptors, and generate discriminative global shape descriptors achieving or matching state-of-the-art results  for non-rigid shape retrieval and classification on diverse  benchmark datasets.

In the future work, we are interested to design an end-to-end learning framework including the raw descriptor extraction on 3D meshes or point clouds. Furthermore, we can also possibly pack the surface second-order pooling stage, SPDM-T stage and  fully connected layer as a block, and add multiple blocks for building a deeper architecture. 

\noindent
\textbf{Acknowledgement.}
This work is supported by National Natural Science Foundation of China under Grants 11622106, 61711530242, 61472313, 11690011, 61721002.

\bibliographystyle{splncs04}
\bibliography{refer}

\begin{thebibliography}{10}
\providecommand{\url}[1]{\texttt{#1}}
\providecommand{\urlprefix}{URL }
\providecommand{\doi}[1]{https://doi.org/#1}

\bibitem{Vincent}
Arsigny, V., Fillard, P., Pennec, X., Ayache, N.: Geometric means in a novel
  vector space structure on symmetric positive-definite matrices. SIAM J.
  Matrix Anal. Appl.  \textbf{29}(1),  328--347 (2007)

\bibitem{MathieuWKS}
Aubry, M., Schlickewei, U., Cremers, D.: The wave kernel signature: A quantum
  mechanical approach to shape analysis. In: ICCV. pp. 1626--1633 (2011)

\bibitem{xiangbai1}
Bai, S., Bai, X., Zhou, Z., Zhang, Z., Tian, Q., Latecki, L.J.: Gift: Towards
  scalable 3d shape retrieval. IEEE Transactions on Multimedia  \textbf{19}(6),
   1257--1271 (2017)

\bibitem{Shapecontext}
Belongie, S., Malik, J., Puzicha, J.: Shape context: A new descriptor for shape
  matching and object recognition. In: NIPS. pp. 831--837 (2001)

\bibitem{MMB4}
Boscaini, D., Masci, J., Rodol{\`a}, E., Bronstein, M.: Learning shape
  correspondence with anisotropic convolutional neural networks. In: NIPS. pp.
  3189--3197 (2016)

\bibitem{MMB1}
Bronstein, M.M., Kokkinos, I.: Scale-invariant heat kernel signatures for
  non-rigid shape recognition. In: CVPR. pp. 1704--1711 (2010)

\bibitem{Carreira}
Carreira, J., Caseiro, R., Batista, J., Sminchisescu, C.: Semantic segmentation
  with second-order pooling. ECCV pp. 430--443 (2012)

\bibitem{Ioannis}
Chiotellis, I., Triebel, R., Windheuser, T., Cremers, D.: Non-rigid 3d shape
  retrieval via large margin nearest neighbor embedding. In: ECCV. pp. 327--342
  (2016)

\bibitem{covariance1}
Cirujeda, P., Mateo, X., Dicente, Y., Binefa, X.: Mcov: A covariance descriptor
  for fusion of texture and shape features in 3d point clouds. In: 3DV. pp.
  551--558 (2015)

\bibitem{dryden2009non-euclidean}
Dryden, I.L., Koloydenko, A., Zhou, D.: Non-euclidean statistics for covariance
  matrices, with applications to diffusion tensor imaging. The Annals of
  Applied Statistics  \textbf{3}(3),  1102--1123 (2009)

\bibitem{elad2003on}
Elad, A., Kimmel, R.: On bending invariant signatures for surfaces. IEEE TPAMI
  \textbf{25}(10),  1285--1295 (2003)

\bibitem{YiFang3}
Fang, Y., Xie, J., Dai, G., Wang, M., Zhu, F., Xu, T., Wong, E.: 3d deep shape
  descriptor. In: CVPR. pp. 2319--2328 (2015)

\bibitem{Furuya2016Deep}
Furuya, T., Ohbuchi, R.: Deep aggregation of local 3d geometric features for 3d
  model retrieval. In: BMVC. pp. 121.1--121.12 (2016)

\bibitem{RMVM}
Gasparetto, A., Torsello, A.: A statistical model of riemannian metric
  variation for deformable shape analysis. In: CVPR. pp. 1219--1228 (2015)

\bibitem{huang2016a}
Huang, Z., Van~Gool, L.: A riemannian network for spd matrix learning. In: AAAI
  (2017)

\bibitem{Catalin1}
Ionescu, C., Carreira, J., Sminchisescu, C.: Iterated second-order label
  sensitive pooling for 3d human pose estimation. In: CVPR. pp. 1661--1668
  (2014)

\bibitem{Catalin2}
Ionescu, C., Vantzos, O., Sminchisescu, C.: Matrix backpropagation for deep
  networks with structured layers. In: ICCV. pp. 2965--2973 (2015)

\bibitem{VLAD}
Jegou, H., Douze, M., Schmid, C., Perez, P.: Aggregating local descriptors into
  a compact image representation. In: CVPR. pp. 3304--3311 (2010)

\bibitem{Andrew}
Johnson, A.E., Hebert, M.: Using spin images for efficient object recognition
  in cluttered 3d scenes. IEEE TPAMI  \textbf{21}(5),  433--449 (1999)

\bibitem{lafon2006data}
Lafon, S., Keller, Y., Coifman, R.R.: Data fusion and multicue data matching by
  diffusion maps. IEEE TPAMI  \textbf{28}(11),  1784--1797 (2006)

\bibitem{BOWVLAD1}
Li, B., Lu, Y., Li, C., et~al.: A comparison of 3d shape retrieval methods
  based on a large-scale benchmark supporting multimodal queries. CVIU
  \textbf{131}(C),  1--27 (2015)

\bibitem{li2015towards}
Li, H., Huang, D., Morvan, J., Wang, Y., Chen, L.: Towards 3d face recognition
  in the real: A registration-free approach using fine-grained matching of 3d
  keypoint descriptors. IJCV  \textbf{113}(2),  128--142 (2015)

\bibitem{Lian2015SHREC}
Lian, Z., Zhang, J., et~al.: Shrec'15 track: Non-rigid 3d shape retrieval.
  Eurographics 3DOR Workshop  (2015)

\bibitem{Tsungyu}
Lin, T.Y., RoyChowdhury, A., Maji, S.: Bilinear cnn models for fine-grained
  visual recognition. In: ICCV. pp. 1449--1457 (2015)

\bibitem{Litman2014Supervised}
Litman, R., Bronstein, A., Bronstein, M., Castellani, U.: Supervised learning
  of bag-of-features shape descriptors using sparse coding. CGF
  \textbf{33}(5),  127--136 (2014)

\bibitem{Luciano2017Deep}
Luciano, L., Hamza, A.B.: Deep learning with geodesic moments for 3d shape
  classification. Pattern Recognition Letters (In press)  (2017)

\bibitem{masci2015geodesic}
Masci, J., Boscaini, D., Bronstein, M., Vandergheynst, P.: Geodesic
  convolutional neural networks on riemannian manifolds. In: ICCV. pp. 832--840
  (2015)

\bibitem{Masoumi2017SHREC}
Masoumi, M., Rodola, E., Cosmo, L.: Shrec'17 track: Deformable shape retrieval
  with missing parts. In: Eurographics 3DOR Workshop (2017)

\bibitem{Ohkita2012Non}
Ohkita, Y., Ohishi, Y., Furuya, T., Ohbuchi, R.: Non-rigid 3d model retrieval
  using set of local statistical features. In: IEEE International Conference on
  Multimedia and Expo Workshops. pp. 593--598 (2012)

\bibitem{Osada2002Shape}
Osada, R., Funkhouser, T., Chazelle, B., Dobkin, D.: Shape distributions. ACM
  TOG  \textbf{21}(4),  807--832 (2002)

\bibitem{Xavier}
Pennec, X., Fillard, P., Ayache, N.: A riemannian framework for tensor
  computing. IJCV  \textbf{66}(1),  41--66 (2006)

\bibitem{SHREC}
Pickup, D., Sun, X., Rosin, P.L., et~al.: Shrec'14 track: shape retrieval of
  non-rigid 3d human models. In: Eurographics 3DOR Workshop (2014)

\bibitem{Mathieu}
Pinkall, U., Polthier, K.: Computing discrete minimal surfaces and their
  conjugates. Experimental Mathematics  \textbf{2}(1),  15--36 (1993)

\bibitem{qsmgpdlps3dcs17}
Qi, C.R., Su, H., Mo, K., Guibas, L.J.: Pointnet: Deep learning on point sets
  for 3d classification and segmentation. Proc. CVPR, IEEE  (2017)

\bibitem{HaoSu1}
Qi, C.R., Su, H., Nie{\ss}ner, M., Dai, A., Yan, M., Guibas, L.J.: Volumetric
  and multi-view cnns for object classification on 3d data. In: CVPR. pp.
  5648--5656 (2016)

\bibitem{qi2017pointnetplusplus}
Qi, C.R., Yi, L., Su, H., Guibas, L.J.: Pointnet++: Deep hierarchical feature
  learning on point sets in a metric space. In: NIPS, pp. 5099--5108 (2017)

\bibitem{reuter2006laplacebeltrami}
Reuter, M., Wolter, F., Peinecke, N.: Laplace-beltrami spectra as 'shape-dna'
  of surfaces and solids. Computer-aided Design  \textbf{38}(4),  342--366
  (2006)

\bibitem{Philip}
Shilane, P., Min, P., Kazhdan, M., Funkhouser, T.: The princeton shape
  benchmark. In: IEEE International Conference on Shape Modeling and
  Applications. pp. 167--178 (2004)

\bibitem{Dirk}
Smeets, D., Keustermans, J., Vandermeulen, D., Suetens, P.: meshsift: Local
  surface features for 3d face recognition under expression variations and
  partial data. CVIU  \textbf{117}(2),  158--169 (2013)

\bibitem{su2015multiview}
Su, H., Maji, S., Kalogerakis, E., Learnedmiller, E.G.: Multi-view
  convolutional neural networks for 3d shape recognition. ICCV pp. 945--953
  (2015)

\bibitem{Jian}
Sun, J., Ovsjanikov, M., Guibas, L.: A concise and provably informative
  multi-scale signature based on heat diffusion. In: CGF. pp. 1383--1392 (2009)

\bibitem{szegedy2015going}
Szegedy, C., Liu, W., Jia, Y., Sermanet, P., Reed, S.E., Anguelov, D., Erhan,
  D., Vanhoucke, V., Rabinovich, A.: Going deeper with convolutions. CVPR
  pp.~1--9 (2015)

\bibitem{Hedi}
Tabia, H., Laga, H., Picard, D., Gosselin, P.H.: Covariance descriptors for 3d
  shape matching and retrieval. In: CVPR. pp. 4185--4192 (2014)

\bibitem{Zhirong}
Wu, Z., Song, S., Khosla, A., Yu, F., Zhang, L., Tang, X., Xiao, J.: 3d
  shapenets: A deep representation for volumetric shapes. In: CVPR. pp.
  1912--1920 (2015)

\bibitem{YiFang1}
Zhu, F., Xie, J., Fang, Y.: Heat diffusion long-short term memory learning for
  3d shape analysis. In: ECCV. pp. 305--321 (2016)

\end{thebibliography}
\end{document}